\documentclass[11pt,a4paper]{article}
\usepackage[hyperref]{emnlp2018}
\usepackage{times}
\usepackage{latexsym}

\usepackage{url}
\usepackage{xcolor}
\usepackage{soul}
\usepackage[utf8]{inputenc}
\usepackage[small]{caption}
\usepackage{mathrsfs}
\usepackage{amsfonts}
\usepackage{indentfirst}
\usepackage{amsmath}
\usepackage{algorithm}
\usepackage{algpseudocode}
\usepackage{amsmath}
\usepackage{graphics}
\usepackage{epsfig}
\usepackage{multirow}
\usepackage{tabularx}
\aclfinalcopy

\title{Keyphrase Generation with Correlation Constraints}

\author{
  Jun Chen$^\dag$, Xiaoming Zhang$^\diamondsuit$,  Yu Wu$^{\dag \heartsuit}$, Zhao Yan$^\ddag$, Zhoujun Li$^\dag$\thanks{~~~Corresponding Author}~~~~~\\
  $^\dag$State Key Lab of Software Development Environment, Beihang University, Beijing, China\\
  $^\diamondsuit$School of Cyber Science and Technology, Beihang University, Beijing, China\\
  	$^\heartsuit$ The author is supported by AdeptMind Scholarship \\
  $^ \ddag$Tencent Lingo Lab, Beijing, China\\
  \{chenjun1994, yolixs, wuyu, lizj\}@buaa.edu.cn \\
  zhaoyan@tencent.com
}

\date{}

\begin{document}
\maketitle
  \begin{abstract}
    
  In this paper, we study automatic keyphrase generation. Although conventional approaches to this task show promising results, they neglect correlation among keyphrases, resulting in duplication and coverage issues. To solve these problems, we propose a new sequence-to-sequence architecture for keyphrase generation named CorrRNN, which captures correlation among multiple keyphrases in two ways. First, we employ a coverage vector to indicate whether the word in the source document has been summarized by previous phrases to improve the coverage for keyphrases. Second, preceding phrases are taken into account to eliminate duplicate phrases and improve result coherence. Experiment results show that our model significantly outperforms the state-of-the-art method on benchmark datasets in terms of both accuracy and diversity.
    
  \end{abstract}
  
  \section{Introduction}

  A keyphrase is a piece of text that is able to summarize a long document, organize contents and highlight important concepts, like "virtual organizations" in Table \ref{tb:issues}.
  It provides readers with a rough understanding of a document without going through its content, and has many potential applications, such as information retrieval, text summarization and document classification.


  \begin{table}[t]
    \scriptsize
    \begin{tabular}{|m{0.946\columnwidth}|}
      \hline
      \hline
      \textbf{Title}:\underline{ Resolving conflict and inconsistency in norm regulated virtual orga-}
      \underline{nizations}.\\
      \textbf{Abstract}: Norm governed virtual organizations define, govern and facilitate coordinated resource sharing and problem solving in societies of agents. With an explicit account of norms, openness in virtual
      organizations can be achieved new components, designed by various parties, can be seamlessly accommodated. We focus on virtual organizations realised as multi agent systems ...\\
      \hline
      \textbf{Ground truth}: virtual organizations; multi agent systems; agent; norm conflict; conflict prohibition; norm inconsistency; ...\\
            \textbf{Predicted keyphrases}: virtual organizations; \textcolor{red}{multi agent systems}; artificial intelligence; inter agent; \textcolor{red}{multi agent}; action delegation; software agents; resource sharing; grid services; \textcolor{red}{agent systems};\\
      \hline
      \hline
    \end{tabular}
    \caption{The example shows the duplication and coverage issues of state-of-the-art model. The phrases in red are 
    duplicate, and the underlined parts in the source document are not covered by the predicted results, while they are summarized by "norm conflict" and "norm inconsistency" in the golden list. }
    \label{tb:issues}
  \end{table}

  Keyphrase can be categorized into present keyphrase which appears in a source document, and absent keyphrase that does not appear in the document.
  Conventional approaches extract important text spans as candidate phrases and rank them as keyphrases \cite{hulth:2003,medelyan:2008,liu:2011,wu2015mining,wang:2016}, that show promising results on the present keyphrases but cannot handle absent keyphrases.  To predict absent keyphrases, generative methods have been proposed by Meng et al. \shortcite{meng:2017}. The approach employs a sequence-to-sequence (Seq2Seq) framework \cite{sutskever:2014} with a copy mechanism \cite{gu:2016} to encourage rare word generation, in which the encoder compresses the text into a dense vector and the decoder generates a phrase with a Recurrent Neural Network (RNN) language model, achieving state-of-the-art performance. Since a document corresponds to multiple keyphrases, the approach divides it into multiple document-keyphrase pairs as training instances. This approach, however, neglects the correlation among target keyphrases since it does not model the one-to-many relationship between the document and keyphrases. Therefore, keyphrase prediction only depends on the source document, and ignores the keyphrases which have been generated. As a consequence, the generated keyphrases suffer from {\bf duplication issue} and {\bf coverage issue}. A duplication issue is defined as at least two phrases expressing the same meaning, hindering readers from obtaining more information from keyphrases. For example, three keyphrases have an identical meaning in Table \ref{tb:issues}, including "multi agent systems", "multi agent" and "agent systems".  A coverage issue means some key points in the document are not covered by the keyphrases, such as "norm conflict" and "norm inconsistency" in Table \ref{tb:issues}.
  

  To mitigate such issues, we mimic human behavior in terms of how to assign keyphrases for an arbitrary document. Given a document in Table \ref{tb:issues}, an annotator will read it and generate keyphrases according to his understanding of the content, like "virtual organizations", "multi agent systems". After that, instead of generating duplicate phrases like "agent systems" and "multi agent", the annotator will review the document and preceding keyphrases, then generate a phrase like "norm conflict" to cover topics that have not been summarized by previous phrases. The iteration does not stop until all of a document's topics are covered by keyphrases.
  
  We propose a new sequence-to-sequence architecture CorrRNN, capable of capturing correlation among keyphrases. Notably, correlation constraints in this paper are defined as keyphrases that should cover all topics in the source document and differ from each other. Specifically, we employ a coverage mechanism \cite{tu:2016} to memorize which parts in the source document have been covered by previous phrases. In this way, the document coverage is modeled explicitly, enabling the generated keyphrases to cover more topics. Furthermore, we propose a review mechanism that considers the previous keyphrases in the generation process, in order to avoid the repetition in the final results. Concretely, the review mechanism explicitly models the correlation between the keyphrases that have been generated and the keyphrase that is going to be generated with a novel architecture. It extends the existing Seq2Seq model and captures the one-to-many relationship in keyphrase generation. Augmented with the coverage mechanism and the review mechanism, CorrRNN does not only inherit the advantages of the Seq2Seq model, but also improves the coverage and diversity in the generation process.
  
  We test our model on three benchmark datasets. The results show that our model outperforms state-of-the art methods by a large margin, demonstrating the effectiveness of the correlation constraints. In addition, our model is better than heuristic rules on improving diversity, since it instills the correlation knowledge to the model in an end-to-end fashion.
  
  Our contributions in this paper are three-fold: (1) the proposal of modeling the one-to-many correlation for keyphrase generation, (2) the proposal of a new architecture CorrRNN  for keyphrase generation, and (3) empirical verification of the effectiveness of CorrRNN on public datasets.

  In the remainder of this paper, we will first review the related work in Section \ref{section:related work}, then we elaborate on the proposed model in Section \ref{section:model}. After that, we list the experiment settings in Section \ref{section:settings}, results and discussion follow in Section \ref{section: results}. Finally, the conclusion and future work in Section \ref{section:conclusion}.
  
  \section{Related Work} \label{section:related work}

  How to assign keyphrases to a long document is a fundamental task, that has been studied intensively in previous works. Existing methods can be categorized into two groups: extraction based and generation based methods. 

    The former group extracts important keyphrases in a document which consists of two phases. The first phase is to construct a set of phrase candidates with heuristic methods, such as extracting important n-grams \cite{hulth:2003,medelyan:2008,hulth:2003,jingboshang:2017} and selecting text chunks with certain postags \cite{liu:2011,wang:2016,le:2016,jialuliu:2015}. The second phase is to rank the candidates with machine learning methods. Specifically, some researchers \cite{frank:1999,witten:1999,hulth:2003,medelyan:2009,sdg-cc:2014} formulate the keyphrase extraction as a supervised classification problem, while others apply unsupervised approaches \cite{mihalcea-tarau:2004,grineva:2009,liu:2009,liu:2010,zhang:2013,bougouin:2013,bougouin:2016} on this task. Besides, Tomokiyo and Hurst \shortcite{tomokiyo-hurst:2003} employ two statistical language models to measure the informativeness for phrases. Liu et al. \shortcite{liu:2011} use a word alignment model to learn translation probabilities between the words in documents and the words in keyphrases, which alleviates the problem of vocabulary gaps.
  
  The latter group, generative methods, assigns keyphrases to a document with natural language generation techniques, and is capable of generating absent keyphrases. Owing to the development of neural networks \cite{bahdanau:2014}, Meng et al. \shortcite{meng:2017} apply an encoder-decoder framework \cite{sutskever:2014} with a copy mechanism \cite{gu:2016} to this task, achieving state-of-the-art performance.

  Our work is a generation based approach. The main difference of our model is that we consider the correlation among keyphrases. Our model proposes a new review mechanism to enhance keyphrase diversity, while employs a coverage mechanism that has proven effective for summarization \cite{see:2017} and machine translation \cite{tu:2016} to guarantee keyphrase coverage. Some previous works on keyword extraction have already exploited the correlation problem with a re-rank strategy \cite{habibi:2013,ni:2012}. In contrast, we model the correlation in an end-to-end fashion.

  \section{\!Keyphrase \!Generation \!with \!Correlation} \label{section:model}
  
  \subsection{Problem Formalization}
  
  Suppose that we have a data set $\mathcal{D} = \{\mathbf{x}_i,\mathbf{p}_i \}_{i=1}^N$, where $\mathbf{x_i}$ is a source text, $\mathbf{p}_{i} = \{p_{i,j}\} _{j=1}^{M_i}$ is the keyphrase set of $\mathbf{x_i}$, and $N$ is the number of documents. Both the source text and target keyphrase are word sequences, donated as $\mathbf{x}_i= (x_{1}^i, x_{2}^i, ..., x_{T}^i) $ and ${p_{i,j}} = (y_1^{i,j}, y_2^{i,j}, ..., y_{L^i}^{i,j}) $ respectively. $T$ and $L^i$ are the length of word sequences of $\mathbf{x_i}$ and ${p_{i,j}}$. Prior work aims to maximize the probability of $\prod_{i=1}^N\prod_{j=1}^{M_i}P(p_{i,j}|\mathbf{x}_i)$, while our model considers keyphrase correlation to address coverage and duplication issues by maximizing the probability of $\prod_{i=1}^N\prod_{j=1}^{M_i}P(p_{i,j}|\mathbf{x}_i, p_{i,l<j})$.
  \newenvironment{sequation}{\begin{equation}\small}{\end{equation}}

  \subsection{Seq2Seq Model with Copy Mechanism}

    \begin{figure*}[ht]
    \centering
    \includegraphics[width=1.\textwidth]{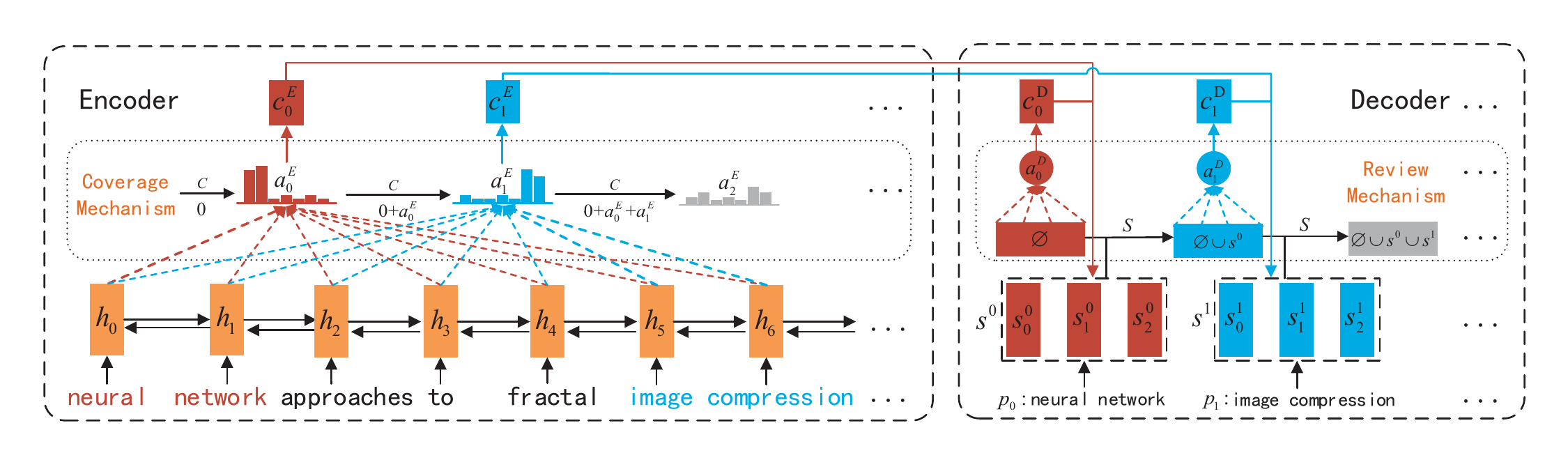}
        \caption{The overall framework structure. Note that $p_i$ indicates a keyphrase (e.g. $p_0=$"neural network"), $s^i$ indicates the hidden state set of phrase $p_i$, coverage vector $C$ and target-side review context $S$ update and transmit along the process of decoding multiple keyphrases.}
    \label{fig:framework}
  \end{figure*}
  
  A Seq2Seq model \cite{sutskever:2014} is employed as backbone in this paper. The encoder converts the variable-length input sequence $x = (x_1, x_2, ..., x_T )$ into a set of hidden representation $h = (h_1, h_2, ..., h_T)$ by iterating along time $t$ with the following equation:
  \begin{sequation} h_t = f(x_t, h_{t-1}) \end{sequation}
  where $f$ is a non-linear function.
  
  Then the context vector $c$ is computed as a weighted sum of hidden representation set $h$ through an attention mechanism \cite{bahdanau:2014}, which next acts as the representation of the whole input $x$ at time step $t$.
  \begin{sequation} c_t = \sum_{j=1}^T\alpha_{tj}h_j \label{eqn:attention} \end{sequation}
  where $\alpha_{tj}$ is a coefficient which measures the match degree between the inputs around position $j$ and the output at position $t$.
  
  With the context vector $c_t$, decoder generates variable-length word sequence step by step, the generative process which is known as a language model:
  \begin{sequation} s_t = f(y_{t-1}, s_{t-1}, c_t) \label{eqn:decoder-rnn} \end{sequation}
  \vspace{-1em}
  \begin{sequation} p_g(y_t|y_{<t}, x) = g(y_{t-1}, s_t, c_t) \label{eqn:decoder-softmax} \end{sequation}
  where $s_t$ denotes the hidden state of the decoder at time $t$. $y_t$ is the predicted word from vocabulary based on the largest probability after $g(.)$.
  
  Unfortunately, pure generative mode cannot generate any keyphrase (e.g. noun, entity) which contains out-of-vocabulary words. Thus we incorporate a copy mechanism \cite{gu:2016} into the encoder-decoder model to predict out-of-vocabulary words by selecting appropriate words from source text. After incorporation, the probability of predicting a new word consists of two parts:
  \begin{sequation} p(y_t|y_{<t},x) = p_g(y_t|y_{<t},x)+p_c(y_t|y_{<t},x) \end{sequation}
  \vspace{-2em}
  \begin{sequation} p_c(y_t|y_{<t},x) = \frac{1}{Z}\sum_{j:x_j=y_t}e^{\sigma(h_j^TW_c)\left[y_{t-1};s_t;c_t\right]}, y_t \in \mathcal{X} \label{eqn:decoder-copy} \end{sequation}
  where $p_g$ and $p_c$ denote the probability of generating and coping. $\mathcal{X}$ is the set of unique words in source sequence ${x}$, $\sigma$ is a non-linear function. $W_c\in\mathbb{R}$ is a learned parameter matrix. $Z$ is the sum for normalization. For more details, please see \cite{gu:2016}.
  
  \subsection{Model Correlation}
  
  Keyphrases should cover more topics and differ from each other, while previous work \cite{meng:2017} ignored this correlation among multiple keyphrases, resulting in duplication and coverage issues. In this part, we focus on capturing the one-to-many correlation to alleviate above issues. On the one hand, we employ a coverage mechanism \cite{tu:2016} that diversifies attention distributions to improve the topic coverage of keyphrases. On the other hand, we propose a review mechanism which makes use of contextual information of previous phrases (already generated) to avoid duplicate generation. For better display of the proposed model, the overall framework is illustrated in Figure \ref{fig:framework} and the detailed process is described in Algorithm \ref{alg::diverseKG}.
  
  \subsubsection{Coverage Mechanism}
  As is well known, multiple keyphrases usually correspond to multiple different positions of the source text (see Figure \ref{fig:framework}), the positions of words that have already been summarized should not be focused again since the attention mechanism automatically focuses on important areas of source text. To overcome the coverage issue, we incorporate a coverage mechanism \cite{tu:2016} into our model, which diversifies the attention distributions of multiple keyphrases to make sure more important areas in the source document are attended and summarized into keyphrases.
  
  Concretely, we maintain a coverage vector $c^t$ which is the sum of attention distributions over all previous decoder time steps. Intuitively, $c^t$ represents the degree of coverage that those words in the source text have received from the attention mechanism so far.
  \begin{sequation} c^t = \sum_{i=0}^{t-1}a^i \end{sequation}
  Note that $c^0$ is a zero vector since no word in source text has be covered.
  
  Later, the coverage vector $c^t$ is an extra input for the attention mechanism, then the source context set $h$ is read and weight averaged into a contextual representation $c_t^E$ by the attention mechanism with a coverage vector, with Eqn.(\ref{eqn:attention}) transforming into Eqn.(\ref{eqn:attention-cov}) as follows:
  \begin{sequation} c_t^E = \sum_{j=1}^T\alpha_{tj}h_j; \ \ \
  \alpha_{tj} = \frac{exp(e_{tj}^E)}{\sum_{k=1}^Texp(e_{tk}^E)};  \nonumber\ \end{sequation}
  \vspace{-1.5em}
  \begin{sequation} e_{tj}^E = v^Ttanh(W_hh_j + W_ss_{t-1} + \mathbf{w_cc_j^t} + b_{attn}) \label{eqn:attention-cov} \end{sequation}
  where $E$ is the encoder and ${\bf w_c}$ is a learned parameter with the same length as $v$.
  
  With the coverage vector, the attention mechanism’s decision for choosing where in source text to focus next is informed by a reminder of its previous decisions, which ensures that the attention mechanism avoids repeatedly attending to the same locations in the source text more easily, thus generated phrases cover more topics in the source document.
  
  \begin{algorithm}[htb]
    \small
    \caption{Training procedure of the proposed model.}
    \label{alg::diverseKG}
    \begin{algorithmic}[1]
      \Require
      The train corpus set $\mathcal{D}$;
      The encoder $\varphi_E$;
      The decoder $\varphi_D$;
      The attention function $attention$;
      \For {each $(X, (P_1, ..., P_M)) \in \mathcal{D}$}
      \State compute source hidden states $H = \varphi_E(X)$;
      \State init target review context $S = \phi$;
      \State init coverage vector $ C = \vec{0}$;
      \For {each $P_i=(y_1^i, y_2^i, ..., y_{T^i}^i) \in (P_1, ..., P_M)$}
      \State init $s_0^i$;
      \For {$t=1$; $t<=T^i$; $t++$}
      \State $c_t^{iE}, a_t^i = attention(H, s_{t-1}^i, C)$ ;
      \State $c_t^{iD} = attention(S, s_{t-1}^i)$ ;
      \State $s_t^i = \varphi_D(y_{t-1}^i, s_{t-1}^i, c_t^{iE}, c_t^{iD})$ ;
      \State $S = S \cup \left\{s_{t-1}^i\right\}$;
      \State $C = C + a_t^i$ ;
      \EndFor
      \State compute $loss$ for $P_i$;
      \EndFor
      \State compute $gradient$ and $update$;
      \EndFor
    \end{algorithmic}
  \end{algorithm}
  
  \subsubsection{Review Mechanism}
  Considering human behavior on assigning keyphrases that review previous phrases to avoid duplicate assignment, we construct a target side review context set which contains contextual information of generated phases. The target context with an attention mechanism can make use of contextual information of generated phrases to help predict the next phrase, which we call the review mechanism.
  
  Like source context $c_t^E$ described above, on the target side, the target context is defined as $s^t = \left\{s_1, s_2, ..., s_{t-1}\right\}$, which is the collection of hidden states before time step $t$. When decoding the word at $t$-th step, $s^t$ is used to inform an extra contextual representation, thus target side attentive contexts are integrated into $c_t^D$:
  \begin{sequation} c_t^D = \sum_{j=1}^{t-1}\beta_{tj}s_j; \ \ \
   \beta_{tj} = \frac{exp(e_{tj}^D)}{\sum_{k=1}^{t-1}exp(e_{tk}^D)}; \nonumber\ \end{sequation}
  \begin{sequation} e_{tj}^D = v^Ttanh(W_hs_j + W_ss_{t-1}) \label{eqn:attention-tgt} \end{sequation}
  \vspace{-1.0em}

  Afterwards, $c_t^D$ is provided as an extra input to derive the hidden state $s_t$ and later the probability distribution for choosing $t$-th word . The target context gets updated consequently as $s^{t+1} = s^t \cup \left\{s_t\right\}$ in the decoding progress.
  \begin{sequation} s_t = f(y_{t-1}, s_{t-1}, c_t^E, \mathbf{c_t^D}) \label{eqn:decoder-tgt-rnn} \end{sequation}
  \vspace{-1em}
  \begin{sequation} p_g(y_t|y_{<t},x) = g(y_{t-1}, s_t, c_t^E, \mathbf{c_t^D}) \label{eqn:decoder-tgt-softmax} \end{sequation}
  \vspace{-1.5em}
  \begin{sequation} p_c(y_t|y_{<t},x) = \frac{1}{Z}\sum_{j:x_j=y_t}e^{\sigma(h_j^TW_c)\left[y_{t-1};s_t;c_t^E;\mathbf{c_t^D}\right]}, y_t \in \mathcal{X}  \label{eqn:decoder-tgt-copy} \end{sequation}
  \vspace{-1.0em}
  
  Eqn.(\ref{eqn:decoder-tgt-rnn}), Eqn.(\ref{eqn:decoder-tgt-softmax}) and Eqn.(\ref{eqn:decoder-tgt-copy}) are transformed from Eqn.(\ref{eqn:decoder-rnn}), Eqn.(\ref{eqn:decoder-softmax}) and Eqn.(\ref{eqn:decoder-copy}) respectively. More mathematical details are displayed below to make Eqn.(\ref{eqn:decoder-tgt-rnn}) more clear:
  \begin{sequation}
  \begin{aligned}
  r_t &= \sigma(W_ry_{t-1}+U_rs_{t-1}+C_r^Ec_t^E+\mathbf{C_r^Dc_t^D});  \\
  z_t &= \sigma(W_zy_{t-1}+U_zs_{t-1}+C_z^Ec_t^E+\mathbf{C_z^Dc_t^D});  \\
  \widetilde{s_t} &= tanh(Wy_{t-1}+U[r_t \circ s_{t-1}]+C^Ec_t^E+\mathbf{C^Dc_t^D});  \\
  s_t &= (1-z_t) \circ s_{t-1} + z_t \circ \widetilde{s_t} \label{eqn:decoder-tgt-rnn-details}
  \end{aligned}
  \end{sequation}
  where $E$, $D$ indicate the encoder and decoder, $W$, $U$, $C$ are learned parameters of the model, $\sigma$ is a $sigmoid$ function, $\circ$ indicates an element-wise product.
  
  With the contextual information of previous phrases, review mechanism ensures next predicted phrase less duplication and topic coherence. So far, we transmit and update the coverage vector and review context along the multi-target phrase decoding process to improve the coverage and diversity of keyphrases. We denote our model with coverage only and review only as CorrRNN$_C$ and CorrRNN$_R$, and empirically compare them in experiments. The objectives are to minimize the negative log-likelihood of the target words, given a data sample with source text $x$ and corresponding phrases set $p = \{p_{i}\} _{i=0}^{M}$, loss is calculated as follows:
  \begin{sequation} loss = - \frac{1}{M}\sum_{i=0}^M\sum_{t=0}^{T_i}log(p(y_t^i|y_{<t}^i, x, p_{j,j<i})) \end{sequation}

  \section{Experiment Settings} \label{section:settings}
  
  \subsection{Implementation Details}
  
  In the preprocessing phase, we follow \cite{meng:2017} to preprocess the text with tokenization, lowercasing, and digit replacement to ensure fairness. Each article consists of one source text and corresponding multiple keyphrases, and the source text is the concatenation of its title and abstract. We set the max number of target phrases to 10 for an article in consideration of the device memory, thus those with more than 10 target phrases are cut into multiple articles. Finally, we have 558830 articles (text-keyphrases pair) for training.
  
  In the training phase, we choose a bidirectional GRU for the encoder and another forward GRU for the decoder. The top 50000 frequent words are chosen as the vocabulary, the dimension of word embeddings is set to 150, the value of embedding is randomly initialized with uniform distribution in [-0.1, 0.1], and the dimension of the hidden layers is set to 300. Adam is adopted to optimize the model with initial learning rate=$10^{-4}$, gradient clipping=0.1 and dropout rate=0.5. The training is stopped once the loss on the validation set stops dropping for several iterations.
  
  In the generation phase, we use beam search to generate multiple phrases. The beam depth is set to 6 and the beam size is set to 200. Source code will be released at \url{https://github.com/nanfeng1101/s2s-kg}.

  \subsection{Datasets}
  Following Meng et al. \shortcite{meng:2017}, we train our model on the \textbf{KP20k} dataset \cite{meng:2017}, which contains articles collected from various online digital libraries. The dataset has 527,830 articles for training and 20000 articles  for validation.
  
  We evaluate our model on three benchmark datasets which are widely adopted in previous works, with the details described below:
  
  \begin{itemize}
    \vspace{-0.5em}
    \item[-]\textbf{NUS} \cite{nguyen-kan:2007}: It contains 211 papers with author-assigned keyphrases, all of which we use as test data.
    \vspace{-1em}
    \item[-]\textbf{Semeval-2010} \cite{kim:2010}: 288 articles are collected from the ACM Digital Library. 100 of them are used for test data and the rest are added to the training set.
    \vspace{-1em}
    \item[-]\textbf{Krapivin} \cite{krapivin:2008}: This dataset contains 2304 papers. The first 400 papers in alphabetical order are used for evaluation and the rest are added to the training set.
    
  \end{itemize}

    \subsection{Baseline Models}

    We compare our model with extractive and generative approaches. The extractive baselines include Tf-idf, TextRank \cite{mihalcea-tarau:2004}, SingleRank \cite{wan-xiao:2008}, ExpandRank \cite{wan-xiao:2008}, TopicRank\footnote{https://github.com/adrien-bougouin/KeyBench} \cite{bougouin:2013}, KEA \cite{witten:1999} and Maui \cite{medelyan:2009}. The generative baselines include RNN and CopyRNN \cite{meng:2017}. In these baselines, the first five are unsupervised and the later four are supervised. We set up all the baselines following optimal settings in \cite{hasan-ng:2010}, \cite{bougouin:2013} and \cite{meng:2017}.

    To demonstrate the effectiveness of end-to-end learning, we compare CorrRNN to CopyRNN with post-processing. In post-precessing, we only keep the first appearence of keyphrase in duplicate predictions, duplication means that a phrase is a substring of another. The baseline can be seen as heuristic rules for improving the diversity, denoted as CopyRNN$_F$.

    \subsection{Evaluation Metrics}

  For a fair comparison, we evaluate the experiment results on present keyphrases and absent keyphrases separately, because extractive methods cannot generate absent keyphrases. Following Meng et al. \shortcite{meng:2017}, we employ ${\rm F1\textrm{-}measure}$ for present keyphrases and ${\rm recall}$ for absent keyphrases. Here, we use \textbf{F1@K} and \textbf{R@K} to denote the F1 and recall score in the top $K$ keyphrases. Note that we use Porter Stemmer for preprocessing to determine whether the two keyphrases are identical.
  
  Furthermore, ${\rm \alpha\textrm{-}NDCG}$, which is widely used to measure the diversity of keyphrase generation \cite{habibi:2013} and information retrieval \cite{clarke:2008}, is adopted to evaluate the diversity of the generative methods, denoted as \textbf{N@K}. $\alpha$ is a trade-off between relevance and diversity in ${\rm \alpha\textrm{-}NDCG}$, which is set to equal weights of $0.5$ according to Habibi and Popescu-Belis \shortcite{habibi:2013}. The higher ${\rm \alpha\textrm{-}NDCG}$ is, the more diverse the results are. We re-implement CopyRNN with the source code\footnote{https://github.com/memray/seq2seq-keyphrase} provided by the authors in order to evaluate it on the ${\rm \alpha\textrm{-}NDCG}$ metric.
    \begin{sequation}
    \alpha\textrm{-}NDCG[k] = \frac{DCG[k]}{DCG'[k]}; \nonumber
    \end{sequation}
    \vspace{-1.5em}
    \begin{sequation}
    DCG[k] = \sum_{j=1}^k G[j]/log_2(j+1); \nonumber
    \end{sequation}
    \vspace{-1em}
    \begin{sequation}
    G[k] = \sum_{i=1}^m J(d_k, i)(1-\alpha)^{r_{i,k-1}} \nonumber
    \end{sequation}
    where $\alpha$ is a parameter, $m$ denotes the number of target phrases, $k$ denotes the number of predicted phrases. $J(d_k, i)=0$ or $1$, which indicates whether the $k$-th predicted phrase is relevant to the $i$-th target phrase, and $r_{i, k-1}$ indicates how many predicted phrases are relevant to the $i$-th target phrase before the $k$-th predicted phrase. Note that relevance here is defined as whether the word set of a keyphrase is a subset of another keyphrase (e.g. "multi agent" vs "multi agent system").

  \section{Results and Analysis} \label{section: results}
  
  \newcommand{\tabincell}[2]{
    \begin{tabular}{@{}#1@{}}#2\end{tabular}
  }

    \subsection{Present Phrase Prediction}
  
  Present phrase prediction is also known as keyphrase extraction in prior studies. We evaluate how well our model performs on this common task. The results are shown in Tables \ref{tb:present-phrases-F1} and \ref{tb:present-phrases-alpha-nDCG}, which list the performance of the top 5 and top 10 results.
  
  In terms of F1-measure, CorrRNN and CopyRNN outperform other baselines by a large margin, indicating the effectiveness of RNN with a copy mechanism. As we consider the correlation among multiple phrases, the overall results of CorrRNN are better than CopyRNN significantly ($t-$test with $p<0.01$). This is mainly because CorrRNN alleviates the duplication and coverage issues in existing methods, with more correct phrases boosted in the top 10 results.
  Heuritic baseline CopyRNN$_F$ is even worse than CopyRNN, indicating that the heuristic rules may hurt the performance of generative approaches. It also proven that it is a better way to model the correlation among keyphrases in an end-to-end fashion.
  
  Regarding ${\rm \alpha\textrm{-}NDCG}$, CorrRNN and its variants surpass other methods, demonstrating that incorporating correlation constraints can improve both relevance and diversity. As the heuristic rules influence the relevance of CopyRNN, CopyRNN$_F$ performs a little better than CopyRNN on the ${\rm \alpha\textrm{-}NDCG}$.

    \begin{table}[htb]
    \scriptsize
    \begin{tabular}{m{0.20\columnwidth}|m{0.20\columnwidth}<{\centering}|m{0.20\columnwidth}<{\centering}|m{0.20\columnwidth}<\centering}
      \hline
      \hline
      {\bf Model} & \tabincell{c}{{\bf NUS} \\ {\bf F1@5 F1@10}} & \tabincell{c}{{\bf SemEval} \\ {\bf F1@5 F1@10}} & \tabincell{c}{{\bf Krapivin} \\ {\bf F1@5 F1@10}}\\
      \hline
      Tf\textrm{-}idf      &      0.136  \ \    0.184   &      0.128  \ \     0.194  &      0.129  \ \     0.160\\
      TextRank    &      0.195  \ \    0.196   &      0.176  \ \     0.187  &      0.189  \ \     0.162\\
      SingleRank  &      0.140  \ \    0.173   &      0.135  \ \     0.176  &      0.189  \ \     0.162\\
      ExpandRank  &      0.132  \ \    0.164   &      0.139  \ \     0.170  &      0.081  \ \     0.126\\
      TopicRank   &      0.115  \ \    0.123   &      0.083  \ \     0.099  &      0.117  \ \     0.112\\
      \hline
      Maui        &      0.249  \ \    0.268   &      0.044  \ \     0.039  &      0.249  \ \     0.216\\
      KEA         &      0.069  \ \    0.084   &      0.025  \ \     0.026  &      0.110  \ \     0.152\\
      RNN         &      0.169  \ \    0.127   &      0.157  \ \     0.124  &      0.135  \ \     0.088\\
      CopyRNN     &      0.334  \ \    0.326   &      0.291  \ \     0.304  &      0.311  \ \     0.266\\
            CopyRNN$_F$  &            0.323  \ \             0.289    &            0.270   \ \             0.270   &             0.293   \ \             0.222 \\
            \hline
      CorrRNN$_C$  &       {\bf 0.361} \ \        {\bf 0.335}   &            0.296   \ \  \underline{0.319}  &             0.311   \ \  \underline{0.273} \\
      CorrRNN$_R$  &            0.354  \ \             0.328    & \underline{0.306}  \ \             0.312   &  \underline{0.314}  \ \             0.270  \\
      CorrRNN      & \underline{0.358} \ \  \underline{0.330}   &       {\bf 0.320}  \ \        {\bf 0.320}  &        {\bf 0.318}  \ \        {\bf 0.278} \\
      \hline
      \hline
    \end{tabular}
    \caption{The ${\rm F1}$ performance on present phrase prediction. From top to bottom, baselines are listed as unsupervised and supervised.}
    \label{tb:present-phrases-F1}
  \end{table}

    \begin{table}[htb]
    \scriptsize
    \begin{tabular}{m{0.20\columnwidth}|m{0.20\columnwidth}<{\centering}|m{0.20\columnwidth}<{\centering}|m{0.20\columnwidth}<{\centering}}
      \hline
      \hline
      {\bf Model}  & \tabincell{c}{{\bf NUS} \\ {\bf N@5 \ \ N@10}} & \tabincell{c}{{\bf SemEval} \\ {\bf N@5 \ \ N@10}} & \tabincell{c}{{\bf Krapivin} \\ {\bf N@5 \ \ N@10}}\\
      \hline
      CopyRNN      &             0.740  \ \    0.713   &      0.682  \ \     0.667  &      0.622  \ \     0.625\\
            CopyRNN$_F$  &             0.743  \ \              0.720    &            0.692   \ \              0.681   &             0.635   \ \         {\bf 0.657} \\
            \hline
      CorrRNN$_C$  &        {\bf 0.781}  \ \  \underline{0.747}   & \underline{0.728}   \ \  \underline{0.694}  &             0.649   \ \              0.642  \\
      CorrRNN$_R$  &             0.770   \ \             0.742    &            0.718   \ \              0.692   &        {\bf 0.669}   \ \        {\bf 0.657} \\
      CorrRNN      &  \underline{0.771}   \ \       {\bf 0.752}   &       {\bf 0.752}  \ \        {\bf 0.720}  &  \underline{0.659}   \ \  \underline{0.647}  \\
      \hline
      \hline
    \end{tabular}
    \caption{The ${\rm \alpha\textrm{-}NDCG}$ performance on present phrase prediction.}
    \label{tb:present-phrases-alpha-nDCG}
  \end{table}

  \subsection{Absent Phrase Prediction}
  
  We evaluate the performance of generative methods within the recall of the top 10 results, which is shown in Table \ref{tb:absent-phrases}. We can see that both CopyRNN and CorrRNN outperform RNN although the improvement is not as much as in present phrase prediction. It indicates that the copy mechanism is very helpful for predicting absent phrases. We can also see that CopyRNN and CorrRNN are comparable in terms of recall, but CorrRNN is better on diversity, proving that our model can address the duplicate issue in keyphrase generation.

    \begin{table}[htb]
    \scriptsize
    \begin{tabular}{m{0.2\columnwidth}|m{0.2\columnwidth}<{\centering}|m{0.2\columnwidth}<{\centering}|m{0.2\columnwidth}<{\centering}}
      \hline
      \hline
      {\bf Model} & \tabincell{c}{{\bf NUS} \\ {\bf R@10 N@10}} & \tabincell{c}{{\bf SemEval} \\ {\bf R@10 N@10}} & \tabincell{c}{{\bf Krapivin} \\ {\bf R@10 N@10}}\\
      \hline
      RNN            &            0.050   \ \ \ \             N/A       & \underline{0.041}  \ \ \ \           N/A    &            0.095   \ \ \ \                  N/A     \\
      CopyRNN        &            0.058  \ \              0.213  &        {\bf 0.043}  \ \             0.228  & \underline{0.113}  \ \                  0.162 \\
            CopyRNN$_F$    &            0.057    \ \            0.216  &        {\bf 0.043}    \ \            0.233   &           0.112   \ \                  0.164 \\
            \hline
      CorrRNN$_C$    &       {\bf 0.064}   \ \             0.215   & \underline{0.041}  \ \             0.231   &       {\bf 0.121}  \ \              {\bf 0.168} \\
      CorrRNN$_R$    &            0.054    \ \  \underline{0.223}  & \underline{0.041}  \ \        {\bf 0.250}  &            0.103   \ \                   0.163 \\
      CorrRNN        & \underline{0.059}   \ \        {\bf 0.229}  & \underline{0.041}  \ \  \underline{0.243}  &            0.108   \ \        \underline{0.166} \\
      \hline
      \hline
    \end{tabular}
    \caption{The ${\rm recall}$ and ${\rm \alpha\textrm{-}NDCG}$ performance on absent phrase prediction.}
    \label{tb:absent-phrases}
  \end{table}
  
\subsection{Generalization Ability}

As described above, CorrRNN performed well on scientific publications. In this part, we construct our experiments on news domain to see if the proposed model works when transferring to a different domain with unfamiliar texts. We adopted the popular news article dataset: DUC-2001 \cite{wan-xiao:2008} for our experiments.  The dataset contains 308 news articles with 2488 manually assigned keyphrases, and each article consists of 740 words on average, which is completely different from the datasets we used above (see Table \ref{tb:dataset-length}). 

\begin{table}[htb]
    \scriptsize
    \begin{tabular}{m{0.2\columnwidth} | m{0.2\columnwidth} | m{0.2\columnwidth} | m{0.2\columnwidth}}
    \hline
    \hline
    {\bf Dataset}  & {\bf Length}    & {\bf Dataset}        & {\bf Length}  \\
        \hline
    NUS            & 219             &  SemEval             & 235          \\
    Krapivin       & 184             & DUC-2001             & 740          \\
    \hline
    \hline
    \end{tabular}
    \caption{The average text length of test datasets.}
    \label{tb:dataset-length}
  \end{table}

We directly applied CorrRNN, which is trained on scientific publications, on predicting phrases for news articles without any adaptive adjustment. Experiment results from \cite{hasan-ng:2010}, \cite{meng:2017} and our experiments are shown in Table \ref{tb:duc2001}, from which we can see that the proposed model CorrRNN can extract a considerable portion of keyphrases correctly from unfamiliar texts. It outperforms TextRank \cite{mihalcea-tarau:2004}, KeyCluster \cite{liu:2009}, TopicRank \cite{bougouin:2013} and CopyRNN \cite{meng:2017}, but it falls behind the other three baselines because the test domain changes. The model should perform better if it is trained on news dataset. 

When transferring to news domain, the vocabulary changes a lot, more unknown words occur, and the correlation also may not applicable, the model can still capture positional and syntactic features within the text to predict phrases despite the different text type and length. The experiment verifies the generalization ability of our model, thus we have good reasons to believe that our model has a great potential to be generalized to more domains after sufficient training.

  \begin{table}[htb]
    \scriptsize
    \begin{tabular}{m{0.20\columnwidth} | m{0.20\columnwidth} | m{0.20\columnwidth} | m{0.20\columnwidth}}
    \hline
    \hline
    {\bf Model}                    & {\bf F1@10}                 & {\bf Model}               & {\bf F1@10}           \\
      \hline
    Tf\textrm{-}idf                & 0.270                       & TopicRank                 & 0.154                 \\
    TextRank                       & 0.097                       & KeyCluster                & 0.140                 \\
    SingleRank                     & 0.256                       & CopyRNN                   & 0.164                 \\
    ExpandRank                     & 0.269                       & CorrRNN                   & 0.173                 \\

    \hline
    \hline
    \end{tabular}
    \caption{Performance on DUC-2001. CopyRNN and CorrRNN are supervised, and they are trained on scientific publications but evaluated on news.}
    \label{tb:duc2001}
  \end{table}

\begin{figure*}[ht]
    \centering
    \includegraphics[width=1.0\textwidth]{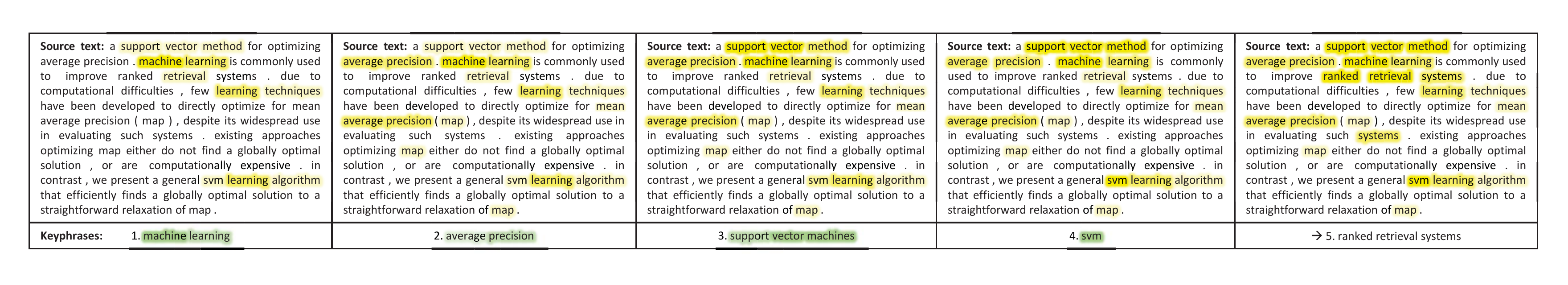}
        \caption{Visualization, deeper shading denotes higher value. Note that yellow shading and green shading indicate coverage vector and review attention respectively.}
    \label{fig:vis}
\end{figure*}

\subsection{Discussion}

    \subsubsection{Model Ablation}
  
  We investigate the effect of coverage mechanism and review mechanism in our model with CorrRNN$_C$ and CorrRNN$_R$ respectively, shown in Tables \ref{tb:present-phrases-F1}, \ref{tb:present-phrases-alpha-nDCG} and \ref{tb:absent-phrases}. It is clear that both the coverage mechanism and review mechanism are helpful for improving the coverage and diversity of predicted phrases. We note the inconsistency of ablate models in our experiments. First, no ablate model achieves the best performance on all of the test datasets, the full model CorrRNN gets better perfermance on present phrase prediction, while CorrRNN$_C$ seems better than the others in absent phrase prediction. As the present phrases are the majority, the full model CorrRNN can achieve best overall performance in actual use. Second, proposed models perform better on dataset NUS and SemEval than Kravipin. This may be due to the difference of assignment quality among test datasets, keyphrase assignments with higher coverage and higher diversity benefiting more from our models.

    \subsubsection{Visualization}

  In Figure \ref{fig:vis}, we visualize the coverage vector and review attention with an example to further clarify how our model works. Due to space limitation, we only visualize top5 phrases in the example, they are already enough to support our analysis. For coverage vector, we can see that source attention transfers along the changes of coverage vector. At the first, only relevant words of "machine learning" are attended. After that, the coverage vector informs attention mechanism to attend other positions instead of repetitive attention, that promotes the generation of later phrases like "average precision". After the last one being generated, it is clear that coverage vector basically covers all topics of the source document, including "machine learning", "mean average precision", "svm" and "ranked retrieval systems". As for the review attention, it's clear that contextual information of all previous phrases are attended for generating the last phrase "ranked retrieval systems", which verifies that the review mechanism helps to alleviate duplication and ensure coherence of results.

    \subsubsection{Comparison with Heuristic Rules}

  We design the baseline CopyRNN$_F$ with post-processing to explore whether heuristic rules can alleviate duplication and coverage issues. It is clear from Tables \ref{tb:present-phrases-F1}, \ref{tb:present-phrases-alpha-nDCG} and \ref{tb:absent-phrases} that the experiment results are negative. Compared to our model CorrRNN, heuristic rules can't address duplication and coverage issues fundamentally. We offer two explanations for these observations. Firstly, heuristic rules can only handle those phrases which have already been generated in results, that shows no help for enabling phrases to cover more topics in source text.  Secondly, although duplication can be reduced by heuristic rules forcibly, the remaining phrases are not guaranteed to be correct, thus it hurts the accuracy badly.

  \subsubsection{Complexity Analysis}
  According to our observations during experiment phase, the CopyRNN \cite{meng:2017} model has 78835750 network parameters, while CorrRNN owns 94886750 parameters because of the incorporation of coverage mechanism (few) and review mechanism (most). However, benefiting from the consideration of one-to-many relationship, and correlation constraints among keyphrases, our model CorrRNN not only achieves better performance but also converges faster than CopyRNN.
  
  \subsubsection{Case Study}

    \begin{table*}[t]
    \scriptsize
    \begin{tabular}{|m{0.98\textwidth}|}
      \hline
      \hline
      \textbf{Title}: Deployment issues of a voip conferencing system in a virtual conferencing environment.\\
      \textbf{Abstract}: Real time services have been supported by and large on circuitswitched networks. Recent trends favour services ported on packet switched networks. For audio conferencing, we need to consider many issues scalability, quality of the conference application, floor control and load on the clients servers to name a few. In this paper, we describe an audio service framework designed to provide a virtual conferencing environment (vce). The system is designed to accommodate a large number of end users speaking at the same time and spread across the internet. The framework is based on conference servers $\langle DIGIT\rangle$, which facilitate the audio handling, while we exploit the sip capabilities for signaling purposes. Client selection is based on a recent quantifier called loudness number that helps mimic a physical face to face conference. We deal with deployment issues of the proposed solution both in terms of scalability and interactivity, while explaining the techniques we use to reduce the traffic. We have implemented a conference server (cs) application on a campus wide network at our institute.\\
      \hline
      \textbf{Present Phrase: } \\
      \textbf{CopyRNN: } deployment; virtual conferencing; real time; distributed systems; \textbf{virtual conferencing environment}; client server; \textbf{conference server}; distributed applications; audio conferencing; floor control;\\
      \textbf{CorrRNN: } \textbf{voip}; virtual conferencing; voip conferencing; audio conferencing; audio service; real time services; real time; distributed systems; \textbf{conference server}; \textbf{virtual conferencing environment}; \\
      \hline
      \textbf{Absent Phrase: } \\
      \textbf{CopyRNN: } quality of service; \underline{distributed conferencing}; virtual environments; \underline{internet conferencing}; \underline{conference conferencing}; load balancing; \underline{packet conferencing}; real time systems; distributed computing; virtual server;\\
      \textbf{CorrRNN: } real time systems; real time voip; voip service; \textbf{real time audio}; wireless networks; conference conferencing; real time communications; packet conferencing; audio communication; quality of service;\\
      \hline
      \hline
    \end{tabular}
    \caption{Top10 generated phrases by CopyRNN and CorrRNN. Phrases in bold are correct, and phrases underlined are duplicate.}
    \label{tb:case}
  \end{table*}

  As shown in Table \ref{tb:case}, we compare the phrases generated by CorrRNN and CopyRNN on an example article. Compared to CopyRNN, CorrRNN generates one more correct present phrase "voip" and one more correct absent phrase "real time audio" respectively, which covers two important topics, while CopyRNN loses these key points. Moreover, four "conferencing (noun)" phrases are generated by CopyRNN, including "distributed conferencing", "internet conferencing", "conference conferencing" and "packet conferencing", which hinders readers from obtaining more information, while CorrRNN only has two.

  \section{Conclusion and Future Work} \label{section:conclusion}
  
  In this paper, we propose a new Seq2Seq architecture that models correlation among multiple keyphrases in an end-to-end fashion by incorporating a coverage mechanism and a review mechanism. Comprehensive empirical studies demonstrate that our model can alleviate duplication and coverage issues effectively and improve diversity and coverage for keyphrase generation. To the best of our knowledge, this is the first use of encoder-decoder model for keyphrase generation in an one-to-many way. Our future work will focus on two areas: investigation on multi-document keyphrase generation, and incorporation of structure or syntax information in keyphrase generation.

  \section*{Acknowledgements}
  We appreciate comments provided by anonymous reviewers. Yu Wu is supported by Microsoft Fellowship Scholarship
  and AdeptMind Scholarship. This work was supported in part by the Natural Science Foundation of China (Grand Nos. U1636211,61672081,61370126), and Beijing Advanced Innovation Center for Imaging Technology (No.BAICIT-2016001) and National Key R\&D Program of China (No.2016QY04W0802).

\bibliography{emnlp2018}

\begin{thebibliography}{34}
\expandafter\ifx\csname natexlab\endcsname\relax\def\natexlab#1{#1}\fi

\bibitem[{Bahdanau et~al.(2014)Bahdanau, Cho, and Bengio}]{bahdanau:2014}
Dzmitry Bahdanau, Kyunghyun Cho, and Yoshua Bengio. 2014.
\newblock Neural machine translation by jointly learning to align and
  translate.
\newblock volume abs/1409.0473.

\bibitem[{Bougouin et~al.(2013)Bougouin, Boudin, and Daille}]{bougouin:2013}
Adrien Bougouin, Florian Boudin, and Beatrice Daille. 2013.
\newblock Topicrank: Graph-based topic ranking for keyphrase extraction.
\newblock In \emph{IJCNLP}, pages 543--551.

\bibitem[{Bougouin et~al.(2016)Bougouin, Boudin, and Daille}]{bougouin:2016}
Adrien Bougouin, Florian Boudin, and Beatrice Daille. 2016.
\newblock Keyphrase annotation with graph co-ranking.
\newblock In \emph{the 26th COLING}, pages 2945--2955.

\bibitem[{Clarke et~al.(2008)Clarke, Kolla, Cormack, Vechtomova, Ashkan,
  Buttcher, and MacKinnon}]{clarke:2008}
Charles L.~A. Clarke, Maheedhar Kolla, Gordon~V. Cormack, Olga Vechtomova, Azin
  Ashkan, Stefan Buttcher, and Ian MacKinnon. 2008.
\newblock Novelty and diversity in information retrieval evaluation.
\newblock pages 659--666. SIGIR.

\bibitem[{Frank et~al.(1999)Frank, Paynter, Witten, Gutwin, and
  Nevill-Manning}]{frank:1999}
Eibe Frank, Gordon~W Paynter, Ian~H Witten, Carl Gutwin, and Craig~G
  Nevill-Manning. 1999.
\newblock Domain-specific keyphrase extraction.

\bibitem[{Gollapalli and Caragea(2014)}]{sdg-cc:2014}
Sujatha~Das Gollapalli and Cornelia Caragea. 2014.
\newblock Extracting keyphrases from research papers using citation networks.
\newblock In \emph{28th AAAI}, pages 1629--1635. AAAI Press.

\bibitem[{Grineva et~al.(2009)Grineva, Grinev, and Lizorkin}]{grineva:2009}
Maria Grineva, Maxim Grinev, and Dmitry Lizorkin. 2009.
\newblock Extracting key terms from noisy and multitheme documents.
\newblock In \emph{18th WWW}, pages 661--670. ACM.

\bibitem[{Gu et~al.(2016)Gu, Lu, Li, and Li}]{gu:2016}
Jiatao Gu, Zhengdong Lu, Hang Li, and Victor O.~K. Li. 2016.
\newblock Incorporating copying mechanism in sequence-to-sequence learning.
\newblock In \emph{Proceedings of the 54th Annual Meeting of the Association
  for Computational Linguistics, {ACL} 2016, August 7-12, 2016, Berlin,
  Germany, Volume 1: Long Papers}.

\bibitem[{Habibi and Popescu-Belis(2013)}]{habibi:2013}
Maryam Habibi and Andrei Popescu-Belis. 2013.
\newblock Diverse keyword extraction from conversations.
\newblock In \emph{ACL (2)}, pages 651--657.

\bibitem[{Hasan and Ng(2010)}]{hasan-ng:2010}
Kazi~Saidul Hasan and Vincent Ng. 2010.
\newblock Conundrums in unsupervised keyphrase extraction: making sense of the
  state-of-the-art.
\newblock In \emph{the 23rd COLING}, pages 365--373. Association for
  Computational Linguistics.

\bibitem[{Hulth(2003)}]{hulth:2003}
Anette Hulth. 2003.
\newblock Improved automatic keyword extraction given more linguistic
  knowledge.
\newblock In \emph{EMNLP 2003}, pages 216--223.

\bibitem[{Kim et~al.(2010)Kim, Medelyan, Kan, and Baldwin}]{kim:2010}
Su~Nam Kim, Olena Medelyan, Min-Yen Kan, and Timothy Baldwin. 2010.
\newblock Semeval-2010 task 5: Automatic keyphrase extraction from scientific
  articles.
\newblock In \emph{5th SemEval}, pages 21--26. Association for Computational
  Linguistics.

\bibitem[{Krapivin et~al.(2008)Krapivin, Autayeu, and Marchese}]{krapivin:2008}
Mikalai Krapivin, Aliaksandr Autayeu, and Maurizio Marchese. 2008.
\newblock Large dataset for keyphrases extraction.
\newblock In \emph{Technical Report DISI-09-055}.

\bibitem[{Le et~al.(2016)Le, Nguyen, and Shimazu}]{le:2016}
Tho Thi~Ngoc Le, Minh~Le Nguyen, and Akira Shimazu. 2016.
\newblock Unsupervised keyphrase extraction: Introducing new kinds of words to
  keyphrases.
\newblock pages 665--671. Springer International Publishing, Cham.

\bibitem[{Liu et~al.(2015)Liu, Shang, Wang, Ren, and Han}]{jialuliu:2015}
Jialu Liu, Jingbo Shang, Chi Wang, Xiang Ren, and Jiawei Han. 2015.
\newblock Mining quality phrases from massive text corpora.
\newblock In \emph{SIGMOD15}.

\bibitem[{Liu et~al.(2011)Liu, Chen, Zheng, and Sun}]{liu:2011}
Zhiyuan Liu, Xinxiong Chen, Yabin Zheng, and Maosong Sun. 2011.
\newblock Automatic keyphrase extraction by bridging vocabulary gap.
\newblock In \emph{15th CoNLL}, pages 135--144. Association for Computational
  Linguistics.

\bibitem[{Liu et~al.(2010)Liu, Huang, Zheng, and Sun}]{liu:2010}
Zhiyuan Liu, Wenyi Huang, Yabin Zheng, and Maosong Sun. 2010.
\newblock Automatic keyphrase extraction via topic decomposition.
\newblock In \emph{EMNLP 2010}, pages 366--376. Association for Computational
  Linguistics.

\bibitem[{Liu et~al.(2009)Liu, Li, Zheng, and Sun}]{liu:2009}
Zhiyuan Liu, Peng Li, Yabin Zheng, and Maosong Sun. 2009.
\newblock Clustering to find exemplar terms for keyphrase extraction.
\newblock In \emph{EMNLP 2009}, volume 1--1, pages 257--266. Association for
  Computational Linguistics.

\bibitem[{Medelyan et~al.(2009)Medelyan, Frank, and Witten}]{medelyan:2009}
Olena Medelyan, Eibe Frank, and Ian~H Witten. 2009.
\newblock Human-competitive tagging using automatic keyphrase extraction.
\newblock In \emph{EMNLP 2009}, volume 3--3, pages 1318--1327. Association for
  Computational Linguistics.

\bibitem[{Medelyan et~al.(2008)Medelyan, Witten, and Milne}]{medelyan:2008}
Olena Medelyan, Ian~H Witten, and David Milne. 2008.
\newblock Topic indexing with wikipedia.
\newblock In \emph{Proceedings of the AAAI WikiAI workshop}, volume~1, pages
  19--24.

\bibitem[{Meng et~al.(2017)Meng, Zhao, Han, He, Brusilovsky, and
  Chi}]{meng:2017}
Rui Meng, Sanqiang Zhao, Shuguang Han, Daqing He, Peter Brusilovsky, and
  Yu~Chi. 2017.
\newblock Deep keyphrase generation.
\newblock In \emph{55th ACL}, volume~1, pages 582--592. Association for
  Computational Linguistics.

\bibitem[{Mihalcea and Tarau(2004)}]{mihalcea-tarau:2004}
Rada Mihalcea and Paul Tarau. 2004.
\newblock Textrank: Bringing order into texts.
\newblock pages 404--411. EMNLP.

\bibitem[{Ni et~al.(2012)Ni, Liu, and Zeng}]{ni:2012}
Weijian Ni, Tong Liu, and Qingtian Zeng. 2012.
\newblock Extracting keyphrase set with high diversity and coverage using
  structural svm.
\newblock In \emph{14th APWeb}, pages 122--133. Springer.

\bibitem[{See et~al.(2017)See, Liu, and Manning}]{see:2017}
Abigail See, Peter~J. Liu, and Christopher~D. Manning. 2017.
\newblock Get to the point: Summarization with pointer-generator networks.
\newblock In \emph{55th ACL}, pages 1073--1083. Association for Computational
  Linguistics.

\bibitem[{Shang et~al.(2017)Shang, Liu, Jiang, Ren, Voss, and
  Han}]{jingboshang:2017}
Jingbo Shang, Jialu Liu, Meng Jiang, Xiang Ren, Clare~R. Voss, and Jiawei Han.
  2017.
\newblock Automated phrase mining from massive text corpora.
\newblock \emph{CoRR}, abs/1702.04457.

\bibitem[{Sutskever et~al.(2014)Sutskever, Vinyals, and Le}]{sutskever:2014}
Ilya Sutskever, Oriol Vinyals, and Quoc~V Le. 2014.
\newblock Sequence to sequence learning with neural networks.
\newblock In \emph{NIPS}, pages 3104--3112.

\bibitem[{Tomokiyo and Hurst(2003)}]{tomokiyo-hurst:2003}
Takashi Tomokiyo and Matthew Hurst. 2003.
\newblock A language model approach to keyphrase extraction.
\newblock In \emph{ACL 2003}, volume~18, pages 33--40. Association for
  Computational Linguistics.

\bibitem[{Tu et~al.(2016)Tu, Lu, Liu, Liu, and Li}]{tu:2016}
Zhaopeng Tu, Zhengdong Lu, Yang Liu, Xiaohua Liu, and Hang Li. 2016.
\newblock Modeling coverage for neural machine translation.

\bibitem[{Vijayakumar et~al.(2016)Vijayakumar, Cogswell, Selvaraju, Sun, Lee,
  Crandall, and Batra}]{nguyen-kan:2007}
Ashwin~K. Vijayakumar, Michael Cogswell, Ramprasaath~R. Selvaraju, Qing Sun,
  Stefan Lee, David~J. Crandall, and Dhruv Batra. 2016.
\newblock Diverse beam search: Decoding diverse solutions from neural sequence
  models.
\newblock volume abs/1610.02424.

\bibitem[{Wan and Xiao(2008)}]{wan-xiao:2008}
Xiaojun Wan and Jianguo Xiao. 2008.
\newblock Single document keyphrase extraction using neighborhood knowledge.
\newblock AAAI.

\bibitem[{Wang et~al.(2016)Wang, Zhao, and Huang}]{wang:2016}
Minmei Wang, Bo~Zhao, and Yihua Huang. 2016.
\newblock Ptr: Phrase-based topical ranking for automatic keyphrase extraction
  in scientific publications.
\newblock pages 120--128. Springer International Publishing, Cham.

\bibitem[{Witten et~al.(1999)Witten, Paynter, Frank, Gutwin, and
  Nevill-Manning}]{witten:1999}
Ian~H Witten, Gordon~W Paynter, Eibe Frank, Carl Gutwin, and Craig~G
  Nevill-Manning. 1999.
\newblock Kea: Practical automatic keyphrase extraction.
\newblock pages 254--255. ACM.

\bibitem[{Wu et~al.(2015)Wu, Wu, Li, and Zhou}]{wu2015mining}
Yu~Wu, Wei Wu, Zhoujun Li, and Ming Zhou. 2015.
\newblock Mining query subtopics from questions in community question
  answering.
\newblock In \emph{AAAI}, pages 339--345.

\bibitem[{Zhang et~al.(2013)Zhang, Huang, and Peng}]{zhang:2013}
Fan Zhang, Lian’en Huang, and Bo~Peng. 2013.
\newblock Wordtopic-multirank : A new method for automatic keyphrase
  extraction.
\newblock In \emph{IJCNLP}, pages 10--18.

\end{thebibliography}
\bibliographystyle{acl_natbib_nourl}

\end{document}